\documentclass[lettersize,journal]{IEEEtran}
\usepackage{amsmath,amsfonts}
\usepackage{algorithm}
\usepackage{array}
\usepackage[caption=false,font=normalsize,labelfont=sf,textfont=sf]{subfig}
\usepackage{textcomp}
\usepackage{stfloats}
\usepackage{url}
\usepackage{verbatim}
\usepackage{graphicx}
\usepackage{cite}

\usepackage{times}
\usepackage{epsfig}
\usepackage{amssymb}
\usepackage[noend]{algpseudocode}
\usepackage{xcolor}
\usepackage{comment}
\newcolumntype{P}[1]{>{\centering\arraybackslash}p{#1}}

\DeclareMathOperator*{\argmin}{argmin}
\DeclareMathOperator*{\argmax}{argmax}

\begin{document}

\title{CBCL-PR: A Cognitively Inspired Model for Class-Incremental Learning in Robotics}

\author{Ali Ayub$^{1,2}$ 
and Alan R. Wagner$^{3}$
\thanks{$^{1}$Department of Electrical and Computer Engineering,
        University of Waterloo, Waterloo, ON N2L3G1, Canada}
\thanks{$^{2}$Concordia Institute for Information Systems Engineering,
        Concordia University, Montreal, QC H3G1M8, Canada
        {\tt\small ali.ayub@concordia.ca}}
\thanks{$^{3}$Department of Aerospace Engineering, The Pennsylvania State University,
       State College, PA 16802, USA
        {\tt\small alan.r.wagner@psu.edu}}%
}

\markboth{IEEE Transactions on Cognitive and Developmental Systems}%
{Shell \MakeLowercase{\textit{Ayub et al.}}: CBCL-PR: A Cognitively Inspired Model for Class-Incremental Learning in Robotics}


\maketitle
\begin{abstract}
\label{sec:Abstract}
For most real-world applications, robots need to adapt and learn continually with limited data in their environments. In this paper, we consider the problem of Few-Shot class Incremental Learning (FSIL), in which an AI agent is required to learn incrementally from a few data samples without forgetting the data it has previously learned. To solve this problem, we present a novel framework inspired by theories of concept learning in the hippocampus and the neocortex. Our framework represents object classes in the form of sets of clusters and stores them in memory. The framework replays data generated by the clusters of the old classes, to avoid forgetting when learning new classes.  
Our approach is evaluated on two object classification datasets resulting in state-of-the-art (SOTA) performance for class-incremental learning and FSIL. We also evaluate our framework for FSIL on a robot demonstrating that the robot can continually learn to classify a large set of household objects with limited human assistance.
\end{abstract}

\begin{IEEEkeywords}
Continual Learning, Catastrophic Forgetting, Few-Shot Learning, Cognitively-Inspired Architectures, HRI.
\end{IEEEkeywords}
\section{Introduction}
\label{sec:introduction}
\noindent
One of the hallmarks of human intelligence is the ability to learn continuously over time. The human brain is flexible enough to learn new tasks quickly (from a few examples), yet stable enough to maintain and recall old tasks over its lifetime. In neuroscience, this is known as the stability-plasticity dilemma \cite{abraham05,mermillod13}. Robots operating in dynamic ever-changing real-world environments face the same dilemma i.e. they must be able to learn about new tasks quickly from a small number of examples without forgetting the previously learned tasks. 

For example, consider a household robot \cite{Matari17, ingen_dynamics_inc_aido_2019} 
tasked with locating and organizing various household objects. The robot must be trained on what items it should organize in a context by its non-expert owner and recognize that the items to be organized might change over time. Further, the robot must learn these items from a few examples because of the limited supervision available from its non-expert owner. 
In this paper, we use concept learning ideas and theories from cognitive science to develop a framework that can allow robots to learn 
novel concepts continually in real-world environments.

Although deep learning has achieved remarkable success in object recognition tasks \cite{He_2016_CVPR}, deep models suffer from \textit{catastrophic forgetting} when learning continually \cite{french19}. Catastrophic forgetting occurs when continually training a system to recognize new object classes, the system drastically forgets the previously learned classes and the overall classification accuracy decreases significantly. Several deep learning techniques have been proposed in recent years to tackle the catastrophic forgetting problem \cite{Rebuffi_2017_CVPR,Castro_2018_ECCV,Wu_2019_CVPR,Ayub_IROS_20,ayub2022neurips}. Most continual learning (also termed as incremental learning) approaches \cite{Rebuffi_2017_CVPR,Castro_2018_ECCV}, however, require data of the old classes to be stored which makes them impractical on robots which usually have a limited amount of on-board memory available. Further, many continual learning approaches \cite{kemker18,Rebuffi_2017_CVPR,Castro_2018_ECCV} require a large number of training examples per class. Robots, on the other hand, typically only have access to limited amount of supervised training data in real-world environments. Thus, it is imperative for robots to learn continually from a few examples per class. We term this problem as \textit{Few-Shot Incremental Learning} (FSIL).

In this paper, we tackle FSIL by developing a novel framework inspired by the ideas of concept learning and replay in the hippocampus and the neocortex \cite{
Sekeres18,Bowman18,Reagh18,Mack18}. Our framework, named \textit{Centroid Based Concept Learning with Pseudo-rehearsal} (CBCL-PR), uses a cognitively-inspired clustering technique to generate multiple clusters of each new class in an increment and store them in the robot's memory. For the classification of objects, CBCL-PR recalls old object classes from its memory and trains a neural network on generated data of the old classes and real data of a new class in an increment. Thus, CBCL-PR avoids catastrophic forgetting without storing the real data of the object classes in its memory. 

We perform extensive evaluations on two benchmark continual learning datasets and on a real robot tasked with learning household objects continually through human assistance. Our results show that CBCL-PR does not suffer from catastrophic forgetting when learning continually, and it significantly outperforms state-of-the-art (SOTA) approaches for continual learning and FSIL. An ablation study of CBCL-PR further demonstrates the contribution of our proposed cognitively-inspired components toward the overall performance of the architecture. Our results suggest that our proposed architecture may allow robots to operate and adapt in real-world environments over the long term.


\section{Related Work}
\label{sec:related_work}


\subsection{Continual Learning}
\noindent Most continual learning techniques proposed in recent years rely on storing a portion of the data from prior classes when learning new classes \cite{Rebuffi_2017_CVPR,Castro_2018_ECCV,Wu_2019_CVPR}. iCaRl \cite{Rebuffi_2017_CVPR} is one of the earliest approaches for class-incremental learning (CiL) that stores data of the old classes. iCaRL uses a regularization term called distillation loss \cite{Hinton15}, which forces the model to keep the labels of the training images of the previous classes to remain the same when learning from new data. iCaRL also uses Nearest Class Mean (NCM) classifier \cite{Mensink13} for classification, instead of the softmax output to mitigate catastrophic forgetting. The NCM classifier uses the distance of the feature vectors of test images from the centroids of each class to predict the labels for the test images. EEIL \cite{Castro_2018_ECCV} and BiC \cite{Wu_2019_CVPR} use similar ideas but improve over iCaRL using end-to-end learning. One of the main limitations of these approaches is that they require storing high-dimensional images of the old classes, which is neither biologically inspired nor practical in situations when the system has limited memory.

To avoid storing real images, various regularization-based techniques have been proposed \cite{kirkpatrick17,Li18,Dhar_2019_CVPR}. These techniques use a regularization loss term to prevent the weights of the model from changing drastically when learning new classes. Another set of approaches uses generative memory to regenerate images of the old classes using generative models \cite{Ayub_ICML_20,Ostapenko_2019_CVPR,ayub2021eec}, and avoid storing the real images of the old classes. Both regularization-based and generative memory-based approaches are more prone to catastrophic forgetting compared to the approaches that store real images of the old classes. One major issue with all CiL approaches is that they require a large amount of training data for each class. Thus, these approaches are not suitable for FSIL. 

As our approach follows the constraints of a class-incremental setup, we mainly review class-incremental learning techniques in this paper. 
A more natural setting of continual learning is \textit{open-ended learning} in which instances of possible classes can be presented in any arbitrary sequence, e.g. \cite{lopes2008open,chauhan2015experimental,kasaei2019local}. We hope to adapt our approach in the future for open-ended learning.

\subsection{Few-Shot Learning}
\noindent Few-shot learning (FSL) has been extensively studied in machine learning in recent years to create models that learn from a few examples per class \cite{
Chen19,Gidaris_2018_CVPR,ren19incfewshot}. Following the general setup of FSL, these approaches first learn a large set of base classes with a large number of training examples to develop a general representation. Then, they learn a small number of classes ($k$) with a few training examples ($n$) and are tested on the test set of the newly learned classes, termed as $n$-shot $k$-way tasks. Some approaches also try to remember the base classes when learning the new classes and are tested on the combined test set of the base and the new classes \cite{Gidaris_2018_CVPR,ren19incfewshot}. However, these approaches do not learn continually for a large number of increments and suffer from catastrophic forgetting. 

\subsection{Few-Shot Incremental Learning}
\noindent For the few-shot incremental learning problem (FSIL) considered in this paper, a model is trained on $n$ examples for $k$ classes in each increment. The model does not have access to the data of the $l$ previously learned classes. After the training process in an increment, the model is evaluated on the test set of all the classes learned ($k+l$). Thus, the model is evaluated on $n$-shot ($k+l$) way tasks in each increment, where $k+l$ keeps increasing in each new increment making the task continually harder. In the past couple of years, there are some approaches that have been proposed for FSIL \cite{Tao_2020_CVPR,Zhang_2021_CVPR}. Tao et al. \cite{Tao_2020_CVPR} use a neural gas representation to preserve the knowledge of the previous classes. Zhang et al. \cite{Zhang_2021_CVPR} use a graph model to preserve past knowledge and use it to adapt new classifiers. However, both of these approaches start with a large number of base classes in the first increment.

\subsection{Robotics Applications of Continual Learning}
\noindent Continual learning has also been applied to real robots in recent years for object learning tasks~\cite{ayub_icra_2021,ayub_icsr_2020}. Turkoglu et al. present a deep learning approach for learning objects continually using mobile robots \cite{Turkoglu18}. They also use prior class data when learning new classes which limit the usefulness of this approach for realistic applications. Similar to Rebuffi et al. \cite{Rebuffi_2017_CVPR}, Deghhan et al. \cite{Dehghan19} present a continual object learning approach that uses the nearest class mean (NCM) classifier \cite{Mensink13} and apply it on a robot manipulator. One important issue with these prior works is that most approaches \cite{Turkoglu18,Dehghan19} continually learn a very small number of object classes (usually 10 or fewer), which is unrealistic. It is also not suitable for testing continual learning approaches, as catastrophic forgetting occurs when learning a large number of classes over a large number of increments (as shown in Section \ref{sec:experiments}). Further, most prior work \cite{Valipour17, Dehghan19} trains and tests the robot on the same objects, which positively skews the results and limits real-world applicability, because in the real-world test data can differ from the training data. In this paper, we present an approach for FSIL that avoids these limitations. 

\subsection{Models of Concept Learning in Cognitive Science}
\noindent Decades of neuroscience research has 
shown that the neural structures associated with concept learning 
are the hippocampus and the neocortex~\cite{
Sekeres18,Bowman18,Reagh18,Mack18}. Based on experimental support from neuroscience, the hippocampus, and the neocortex use several mechanisms to learn and store new memories \cite{Mack18,Sekeres18,Reagh18,genzel17}. The theoretical models proposed in this research suggest that whenever a new episode (stimulus/event) is encountered by the hippocampus it first extracts feature information about the episode. Then the difference between the feature map of the episode to previously learned concepts is calculated. This difference is known as \textit{memory-based prediction error}. When the memory-based prediction error is large, the hippocampus performs \textit{pattern separation} by creating a new, distinct concept for the episode. When the memory-based prediction error is small, \textit{memory integration} is performed by updating an existing concept to incorporate the new episode.  

Evidence from neuroscience also suggests that long-term memories are recalled by the neocortex for memory 
replay during REM sleep \cite{Gais07,Robins95}. Using a combination of these processes, the mammalian brain continues to develop and store memories over time without catastrophic forgetting. CBCL-PR is inspired by these mechanisms of concept learning and memory replay in the mammalian brain for class-incremental learning in robots.

\begin{figure*}
\centering
\includegraphics[width=0.9\linewidth]{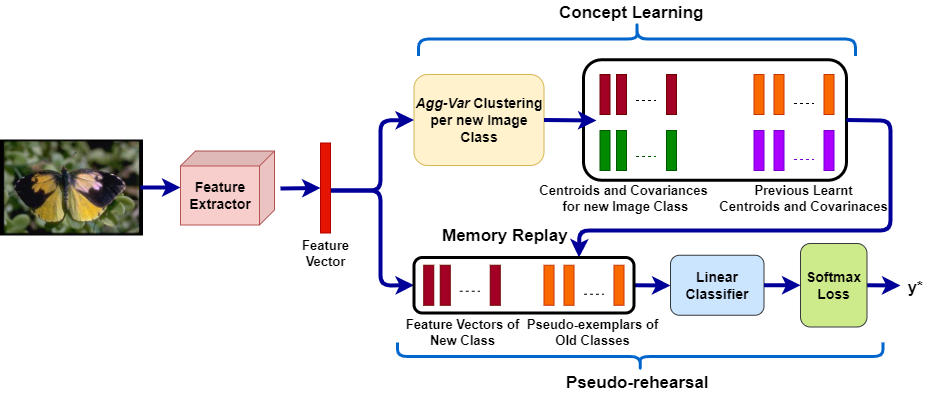}
\caption{\small Complete architecture of CBCL-PR. The model extracts features for the training images of a class and generates clusters using \textit{Agg-Var} clustering. Pseudo-exemplars of the old classes are then generated which are used with the feature vectors of the new class to train a shallow classifier.}
\label{fig:framework_cbcl_pr}
\end{figure*}


\section{Centroid-Based Concept Learning with Pseudo-rehearsal}
\label{sec:methodology}
\noindent CBCL-PR consists of two major phases, 1) concept learning phase and 2) pseudo-rehearsal phase, for continual learning of object classes. We use a pre-trained convolutional neural network (CNN) on image data of object classes to generate feature vectors which are then fed to CBCL-PR. Figure \ref{fig:framework_cbcl_pr} shows our complete CBCL-PR architecture. 

\subsection{\textit{Agg-Var} Clustering}
\label{sec:agg_var_cbcl_pr}
\noindent
The learning phase of CBCL-PR consists of a clustering approach (termed as \textit{Agg-Var} clustering) inspired by the concept learning models in the hippocampus and the neocortex. For the continual learning setting, when CBCL-PR receives the training data of a new image class $1 \leq y \leq N$ in a new increment, \textit{Agg-Var} clustering is applied on the feature vectors $\{x_1^y, x_2^y, ..., x_{N_y}^y\}$ of the class. The process begins by initializing a new cluster for the class $y$, with the centroid $c_1^y$ equated to the first feature vector $x_1^y$ of the class. Next, each feature vector $x_i^y$ is compared to all the clusters of class $y$ by computing the Euclidean distance between the feature vector $x_i^y$ and the centroids of the clusters of class $y$. If the distance $dist(x_i^y,c_{min}^y)$ between the feature vector $x_i^y$ and the closest centroid $c_{min}^y$ is lower than the distance threshold $D$ (a hyperparameter), $x_i^y$ is integrated into the corresponding cluster (similar to \textit{memory integration}), and the closest centroid $c_{min}^y$ is updated by taking a weighted mean of the centroid and the feature vector:

\begin{equation}\label{equation1}
c_{min}^y = \frac{w_c \times c_{min}^y + x_i^y}{w_c+1}
\end{equation}

\noindent where, $w_c$ is the number of feature vectors already clustered in the cluster represented by the centroid $c_{min}^y$. If, on the other hand, distance $dist(x_i^y, c_{min}^y)$ is greater than the distance threshold $D$, a new cluster is initialized for class $y$, with the centroid of the cluster equated to the feature vector $x_i^y$ (similar to \textit{pattern separation}). 

\begin{figure*}
\centering
\includegraphics[width=0.85\linewidth]{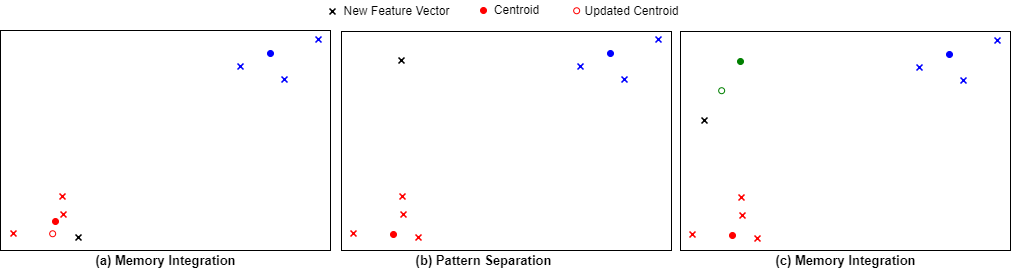}
\caption{\small The complete process of \textit{Agg-Var} clustering. (a) New feature vector closest to one of the previous centroids of the same class, is integrated in the cluster and a new centroid is generated. (b) In the next increment, a new feature vector is far from the two centroids, so a new cluster is generated with new feature vector as the centroid. (c) Another new feature vector for the class is integrated in the previous cluster and a new centroid is generated.}
\label{fig:agg_var}
\end{figure*}

This process is applied on the all feature vectors of the new class $y$ which results in a set of clusters $N^*_y$ for the class\footnote{$N^*_y$ can be different for different classes and is unknown beforehand. The total number of clusters for each class is dependant on the similarity among the images (intra-class variance) of the class, thus it is determined by the distance threshold $D$.}. 
After finding the clusters for all the feature vectors, \textit{Agg-Var} clustering also computes the covariance matrices for the clusters of class $y$ using the corresponding feature vectors. Thus, the clusters for class $y$ are represented by a set of centroids $C^y = \{c_1^y, ..., c_{N^*_y}^y\}$ and covariance matrices $\sum^y = \{\sigma_1^y,...,\sigma_{N^*_y}^y\}$, both of which are stored in the system's memory. Note that after the centroids and covariance matrices for the new class are computed and stored in memory, the original feature vectors are discarded. An intuitive explanation of \textit{Agg-Var} clustering is illustrated in Figure \ref{fig:agg_var}, and a formal description of the complete process of \textit{Agg-Var} clustering in a single increment is described in Algorithm 1.

\begin{algorithm}
\caption{CBCL-PR: \textit{Agg-Var} Clustering}
\begin{flushleft}
        \textbf{Input:} $X=\{X^1,...,X^t\}$\Comment{feature vector sets of $t$ classes}\\
        \textbf{require:} $D$\Comment{distance threshold}\\
        \textbf{Output:} $C = \{C^1,...,C^t\}$\Comment{Centroid sets for $t$ classes}\\
        \textbf{Output:} $\sum = \{\sum^1,...,\sum^t\}$\Comment{Covariance matrices sets for $t$ classes}\\
\end{flushleft}
\begin{algorithmic}[1]
\For {$y=1$; $j \leq t$} 
\State $C^y\leftarrow\{x_1^y\}$\Comment{initialize centroids for each class}
\EndFor
\For {$y=1$; $y\leq t$}
\For {$i=2$; $i \leq N_y$}
\State $d_{min} \leftarrow \min_{l=1,..,N^*_y} dist(c^y_{l},x^y_{i})$\Comment{distance to closest centroid}
\State $i_{min} \leftarrow \argmin_{l=1,..,N^*_y} dist(c^y_{l},x^y_{i})$\Comment{index of the closest centroid}
\State \textbf{Set} $w^y_{i_{min}}$ to be the number of images clustered 
\State in the $i_{min}$th cluster of class $y$
\State Assign $i_{min}$ as the cluster label to $x_i^y$
\If {$d_{min}<D$}
\State $c^y_{i_{min}} \leftarrow \frac{w^y_{i_{min}} \times c^y_{i_{min}} + x^y_i}{w^y_{i_{min}}+1}$\Comment{see equation (\ref{equation1})}
\Else
\State $C^y.append(x^y_i)$\Comment{new centroid for class $y$}
\EndIf
\EndFor
\For {$i=1$;$i\leq N^*_y$}
\State $X^{y*}_i = \{x_k^y | x_k^y \in c_i^y \} \forall k \in \{1,...,N_y\}$\Comment{get all feature vectors belonging to $i$th cluster of class $y$}
\State $\sum^y.append(cov(X^{y*}_i))$\Comment{add covariance matrix for the feature vectors in cluster $i$ to the set $\sum^y$}
\EndFor
\EndFor

\end{algorithmic}
\end{algorithm}

\subsection{Pseudo-rehearsal}
\label{sec:pseudorehearsal_cbcl_pr}
\noindent 

\noindent Inspired by memory replay during the sleep phase of the mammalian brain, CBCL-PR generates pseudo-exemplars using the clusters of the old classes to train a classifier. The process is known as pseudo-rehearsal (also called intrinsic replay) \cite{Robins95}.

Let $M^y = \{m_1^y,...m_i^y,...,m_{N^*_y}^y\}$ represent the number of images clustered in all the $N^*_y$ clusters of class $y$. During pseudo-rehearsal, each centroid $c_i^y$ and covariance matrix $\sigma_i^y$ pair of class $y$ is used to initialize a multivariate Gaussian distribution. Each of the Gaussian distributions $\mathcal{N}(c_i^y,\sigma_i^y)$ are sampled to generate $m_i^y$ number of pseudo-exemplars for class $y$. For a balanced training set, the model generates a total of $N_y$ pseudo-exemplars after sampling all the Gaussian distributions, where $N_y$ is the original number of feature vectors of class $y$. 

In a new increment, the pseudo-exemplars of the previous classes are mixed with the real feature vectors of the new classes in the increment. The mixture is used to train a linear classifier using back-propagation. During the test phase, the model simply passes an unlabeled feature vector $x$ through the classifier, to predict the label $y^*$ as: 

\begin{equation}
    y^* = \argmax_y (w_y^Tx)
\end{equation}

\noindent where, $w_y^T$ is the weight vector of the linear classifier that transforms the feature vector $x$ into probability values. Algorithm 2 formally describes the complete process of pseudo-rehearsal.

\begin{figure}[t]
\centering
\includegraphics[width=1.0\linewidth]{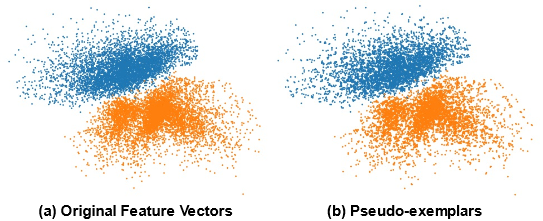}
\caption{\small Comparison of the original feature space (11379 feature vectors) with the pseudo-exemplars generated by 1000 clusters for 2 classes in the MNIST dataset.}
\label{fig:mnist_example}
\end{figure}

To show an illustration of pseudo-rehearsal, we performed a small experiment on 2 classes in the MNIST dataset \cite{Lechun98}. Figure \ref{fig:mnist_example} shows the original feature vectors and pseudo-exemplars for the 2 classes in the MNIST dataset. The 2D space covered by the pseudo-exemplars looks almost the same as the original feature vectors. This shows that CBCL-PR can learn the complex distributions of the image data using clusters, and it can use the clusters to regenerate pseudo-exemplars that are significantly similar to the original feature vectors. Finally, note that an object class can contain multiple clusters representing clustered groups of different feature vectors of the class.

\begin{algorithm}
\caption{CBCL-PR: Pseudo-rehearsal}
\begin{flushleft}
        \textbf{require:} $X^t = \{x_1^t, x_2^t, ..., x_{N_t}^t\}$\Comment{training samples of new class $t$}\\
        \textbf{require:} $C = \{C^1,...,C^{t-1}\}$\Comment{centroid sets for old classes}\\
        \textbf{require:} $\sum = \{\sum^1,..., \sum^{t-1}\}$\Comment{covariance matrices for old classes}\\
        \textbf{require:} $M = \{M^1,...,M^{t-1}\}$\Comment{Number of samples clustered in clusters of old classes}\\
        \textbf{require:} $D$ \Comment{Linear classifier}
\end{flushleft}
\begin{algorithmic}[1]
\For{$y=1$;$y<t$}
\State $X^{*y} = \{ \}$\Comment{empty variable to temporarily store pseudo-exemplars of class $y$}
\For{$i=1$;$i \leq N^*_y$}
\State $X^{*y}.append(multivariate\_normal(c_i^y,\sigma_i^y,m_i^y))$\Comment{pseudo-exemplars using each centroid and covariance matrix pair of class $y$}
\EndFor
\EndFor
\State $X^* = \{X^{*1},X^{*2},...,X^{*t-1},X^t\}$\Comment{Mixture of pseudo-exemplars and new feature vectors}
\State Train parameters $\theta$ of $D$ with $X^*$ using back-propagation with cross-entropy loss:
\State $l(\theta) = - \Sigma_{y=1}^t [\Sigma_{i=1}^{N_y} \delta_{y=y_i^*}{\rm{log}}(p(x_i^y)) + \delta_{y \neq y_i^*}{\rm{log}}(1-p(x_i^y))]$

\end{algorithmic}
\end{algorithm}

\subsection{CBCL-PR for FSIL}
\noindent The sections above describe our method for a general continual learning setting. For few-shot incremental learning (FSIL), the \textit{Agg-Var} clustering part of CBCL-PR remains unchanged and, for each set of new classes, a set of centroids and covariance matrices are generated and stored in memory. The pseudo-rehearsal phase changes slightly for FSIL. Since each class only has a small set of training examples (usually 5 or 10), we do not use the real feature vectors of a new class when training the linear classifier. Instead we generate a larger number of pseudo-exemplars (usually 40) for all the classes (old and new) in each increment and train the linear classifier using the generated pseudo-exemplars.

\section{Prediction Time Analysis of CBCL}
\label{sec:analysis_cbcl}
\noindent Our previously proposed vanilla CBCL method \cite{Ayub20} uses a weighted voting scheme for classification of unlabeled images. However, prediction time for the weighted voting scheme increases as the number of clusters increases. The prediction time complexity of CBCL's weighted voting scheme for classification (see \cite{Ayub20} for details) is determined with respect to the number of clusters (centroids) learned so far ($N_C$). Thus:

\begin{itemize}
    \item The time required to find the distance of the test feature vector from the complete set of centroids learned so far is $O(N_C)$;
    \item Plus, the time required to find top $n$ closest centroids to the test feature vector, $O(N_C{\rm{log}}(N_C))$;
    \item Plus, the time required to update the prediction weights of all $t \leq N_C$ classes learned so far using the distances of the test feature vector from the $n$ closest centroids, $O(t)$;
    \item Plus, the time required to find the class with the highest prediction weight, $O(t)$.
\end{itemize}

Thus, the time complexity ($T(N_C)$) as a function of the number of centroids learned so far is:

\begin{equation} \label{eq_time_complexity}
    \begin{split}
    T(N_C) &= O(N_C) + O(N_C{\rm{log}}(N_C)) + O(t) + O(t)\\
    & = O(N_C{\rm{log}}(N_C))
    \end{split}
\end{equation}

Equation \ref{eq_time_complexity} shows that time complexity of the weighted voting scheme for classification is directly proportional to the number of clusters $N_C$. Thus, as the model learns more classes continually, the total number of clusters increases which in turn increases the prediction time of the model. Section \ref{sec:experiments} shows a comparison of the prediction times of CBCL and CBCL-PR on the Caltech-101 dataset when incrementally learning 10 classes per increment which confirms our analysis.  
\section{Experiments}
\label{sec:experiments}
\noindent In this section, we first describe the datasets and implementation details for the experiments. We then compare CBCL-PR to state-of-the-art (SOTA) approaches and the batch learning baseline for continual learning, and 5-shot and 10-shot incremental learning. 
Finally, we evaluate CBCL-PR for few-shot incremental learning of household objects on the Baxter robot. Our code is available at \url{https://github.com/aliayub7/CBCL-PR}.


\begin{table}[t]
\centering
\small
\caption{Total Number of classes, number of training and test images per class and number of classes per increment for the CIFAR-100 and Caltech-101 datasets.}
\begin{tabular}{ p{2.2cm}p{1.8cm}p{1.8cm} }
     \hline
    \textbf{Dataset} & \textbf{CIFAR-100} 
    & \textbf{Caltech-101} \\
     \hline
    \# classes & 100 
    & 100\\
     \hline
    \# training images &  500 
    & 80\% of data\\
    \# testing images &  100 
    & 20\% of data\\
    \# classes/batch & 2, 5, 10, 20 
    & 10\\
 \hline
 \end{tabular}
 
 \label{tab:Datasets}
 \end{table}

\subsection{Datasets}
\noindent
We evaluated CBCL-PR on two object classification datasets that have been used previously to test class-incremental learning techniques: CIFAR-100 \cite{Krizhevsky09} and Caltech-101 \cite{fei-fei06} 
. 
CIFAR-100 contains 60,000 images of resolution $32\times32$ belonging to 100 object classes. Each image class is composed of 500 training and 100 test images. Caltech-101 consists of 101 object classes with 8,677 total images and 40 to 800 images per class. The statistical details for both of the datasets are described in Table \ref{tab:Datasets}. These settings are in accordance with the previous works \cite{Dhar_2019_CVPR,Ayub20,Rebuffi_2017_CVPR} for a fair comparison. For FSIL experiments, the training images for each class in Table \ref{tab:Datasets} were changed to 5 and 10 for 5-shot and 10-shot incremental learning experiments, respectively, while all the other statistics were kept the same. We report top-1 accuracy in each increment and the average of the accuracies achieved over all increments (termed as average incremental accuracy).
We performed all of our experiments 10 times with different random seeds to randomize the order of the classes in different experiments. Average and standard deviation of the accuracy achieved over the 10 runs are reported.  

\begin{figure*}
\centering
\includegraphics[width=0.85\linewidth]{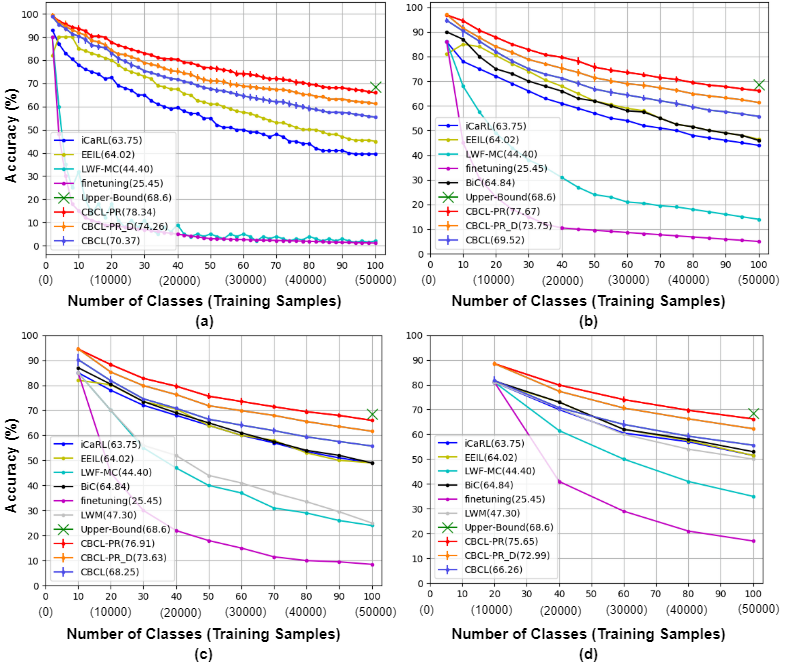}
\caption{\small Accuracy curves on the CIFAR-100 dataset for all the models trained with (a) 2, (b) 5, (c) 10, (d) 20 classes per increment. Mean and standard deviation of the accuracies over 10 runs are reported for our method. Average incremental accuracy is shown in parenthesis for each method. CBCL-PR\_D shows our method with the diagonal entries for the covariance matrices. (Best viewed in color)}
\label{fig:cifar}
\end{figure*}

\subsection{Implementation Details}
\label{sec:implementation_details}
\noindent
We used the Pytorch framework \cite{torch19} to implement and train the neural network models. We used the ResNet-18 and ResNet-34 \cite{He_2016_CVPR} models pre-trained on the ImageNet dataset \cite{Russakovsky15} as feature extractors for the Caltech-101 and CIFAR-100 datasets, respectively. These model choices are consistent with \cite{Dhar_2019_CVPR,Ayub20,Rebuffi_2017_CVPR} for a fair comparison. For continual learning, when our method uses all the training examples for each class, storing the covariance matrices can require a large amount of memory. Hence, we also show results using our method when it only stores the diagonal entries of the covariance matrices which reduces the memory requirement by $\sim$99\%, although the decrease in accuracy is only $\sim$5\%. For FSIL, as there are only a small number of clusters per class, we stored the complete covariance matrices per class without requiring a huge memory budget.

As mentioned in Section \ref{sec:introduction}, since none of the prior incremental learning approaches are applicable on FSIL, we compare our method to CBCL and a few-shot batch learning baseline (FLB) \cite{Chen19}. FLB also uses a pre-trained feature extractor (ResNet) to get the feature vectors for the training images. The feature vectors are then used to train a shallow linear classifier \cite{Chen19}. FLB is not designed for continual learning and will suffer from catastrophic forgetting if trained continually. Therefore, we train FLB using the training data of all the old and new classes in each increment. In other words, FLB is the batch learning baseline and, as such, should have a considerable advantage in terms of classification accuracy over continual learning approaches.

For CBCL-PR, the distance threshold ($D$) for Agg-Var clustering was set to 20.0 in all the experiments. For the linear classifier used for classification, we use a linear layer of the same size as the feature vectors generated by the ResNet. 
Both FLB and the linear classifier in CBCL-PR were trained for 25 epochs using the cross-entropy loss optimized with stochastic gradient descent (with 0.9 as momentum) in the first increment and we kept on increasing the number of epochs by 2 for each new increment. A fixed learning rate of 0.001 with minibatches of size 8 were used for FLB and a fixed learning rate of 0.01 and minibatches of size 64 were used for the linear classifier in CBCL-PR. 


\subsection{Results on the CIFAR-100 Dataset}
\label{sec:Cifar_cbcl_pr}
\noindent We compare CBCL-PR to seven state-of-the-art (SOTA) methods on CIFAR-100 dataset: Finetuning (FT), LWM \cite{Dhar_2019_CVPR}, LWF-MC~\cite{Rebuffi_2017_CVPR}, iCaRL~\cite{Rebuffi_2017_CVPR}, EEIL~\cite{Castro_2018_ECCV}, BiC~\cite{Wu_2019_CVPR} and CBCL \cite{Ayub20}. All of these methods have been introduced in Section \ref{sec:related_work}. We also compare our classification accuracy to a batch learning upperbound with a ResNet-34 trained on the complete CIFAR-100 dataset in a single batch.

Figure~\ref{fig:cifar} shows a comparison of CBCL-PR and CBCL-PR\_D (CBCL-PR with only diagonal entries of the covariance matrices) to other methods for class-incremental learning experiments with 2, 5, 10 and 20 classes per increment. CBCL-PR outperforms all seven approaches (including CBCL) on all increment settings by significant margins. Note that iCaRL, EEIL and BiC store and use the real images of the old classes, requiring more memory than CBCL-PR, but they perform significantly lower accuracy than CBCL-PR. Even CBCL-PR\_D outperforms all seven approaches while using significantly lesser memory. The accuracy of the other methods deteriorates quickly compared to CBCL-PR, which shows the ability of our model to retain most of the past knowledge when learning incrementally. While other approaches produce lower final accuracy when learning with a smaller number of classes per increment (Figure \ref{fig:cifar} (a)), CBCL-PR and CBCL-PR\_D produce the same final accuracy for all increment settings. It should also be noted that when using the complete covariance matrices our method achieves an accuracy that is only $\sim$1\% lower than the batch learning upperbound. These results show that, unlike batch learning systems, our approach has the advantage of learning continually without 
storing all the original data, while also achieving performance similar to the batch learning systems. 

For this experiment, the maximum number of clusters that were allowed to be stored by our model was 1600, which requires a total memory of 1.63 MB. However, other approaches \cite{Rebuffi_2017_CVPR,Castro_2018_ECCV,Wu_2019_CVPR} usually store $\sim$2000 images for the previously learned classes, which requires 17.6 MB. Thus, our approach is computationally more efficient which allows for application on real-world systems with limited computational resources. 

\begin{table}[t]
\centering
\caption{Results on FSIL settings in terms of average incremental accuracy (\%) on the CIFAR-100 dataset.}
\begin{tabular}{ |P{1.9cm}|P{0.9cm}|P{0.8cm}|P{0.8cm}|P{0.8cm}|P{0.8cm}| }
    \hline
    \multicolumn{2}{|c}{} & \multicolumn{4}{|c|}{\textbf{Classes per increment}}\\
     \hline
    \textbf{Methods} &\textbf{k-Shot} & \textbf{2} & \textbf{5} & \textbf{10} &\textbf{20} \\
    \hline
    FLB & 5 & 41.1 & 39.9 & 41.3 & 44.4 \\
    \hline
    CBCL & 5 & 56.9 & 55.6 & 54.7 & 53.8 \\
    \hline
    \textbf{CBCL-PR} & 5 & \textbf{57.9} & \textbf{57.2} & \textbf{57.0} & \textbf{53.9} \\
    \hline
    FLB & 10 & 53.5 & 52.4 & 55.1 & 55.6 \\
    \hline
    CBCL & 10 & 61.9 & 61.4 & 61.3 & 60.7\\
    \hline
    \textbf{CBCL-PR} & 10 & \textbf{64.5} & \textbf{63.8} & \textbf{62.5} & \textbf{60.7} \\
 \hline
 \end{tabular}
 \label{tab:average_increment_cbcl_pr}
 \end{table}


\subsubsection{FSIL Results}Table \ref{tab:average_increment_cbcl_pr} shows a comparison of CBCL-PR with CBCL and FLB for the few-shot incremental learning experiments in terms of average incremental accuracy. For both 5-shot and 10-shot incremental learning and all four increment settings, CBCL-PR outperforms FLB by $\sim$\textbf{11-17\%}. Note that FLB is the batch learning baseline and has the advantage of training on the data of all the classes, still CBCL-PR outperforms FLB significantly. Similar to continual learning results, CBCL-PR produces higher accuracy than CBCL for all FSIL settings. Also, note that CBCL-PR outperforms FLB with the largest margin for 5-shot incremental learning and using 2 classes per increment. These results show that our method is best suitable for real-world situations with limited data in a single increment.     
 
We also compare CBCL-PR with other SOTA approaches for FSIL on the CIFAR-100 dataset. All of these approaches have been introduced in Section \ref{sec:related_work}. For a fair comparison with these approaches, we use the training and test settings proposed in TOPIC \cite{Tao_2020_CVPR} and CEC \cite{Zhang_2021_CVPR}, where a large number of CIFAR-100 classes are learned in the first increment and the rest of the classes are learned in 10 increments. Table \ref{tab:fsil_sota} compares the average incremental accuracy of all the approaches with CBCL-PR for FSIL. NCM, iCaRL and EEIL produce significantly lower accuracy compared to CBCL-PR, as these methods are not specifically designed to solve FSIL. Compared to the other two methods (TOPIC and CEC) that are specifically designed to solve FSIL, CBCL-PR still outperforms them by margins of 18.9\% and 2.6\%, respectively.  

 \begin{table}[t]
 \caption{Comparison of CBCL-PR with SOTA methods for FSIL on the CIFAR-100 dataset.} 
 \label{tab:fsil_sota}
\centering
\begin{tabular}{ |P{3cm}|P{3.0cm}|}
     \hline
    \textbf{Methods} & \textbf{Accuracy (\%)} \\
     \hline
    NCM \cite{Mensink13} & 36.9\\
    \hline
    iCaRL \cite{Rebuffi_2017_CVPR} & 35.5\\
    \hline
    EEIL \cite{Castro_2018_ECCV} & 37.5 \\
     \hline
    TOPIC \cite{Tao_2020_CVPR} & 43.1\\
    \hline
    CEC \cite{Zhang_2021_CVPR} & 59.4\\
    \hline
    \textbf{CBCL-PR} & \textbf{62.0} \\
 \hline
 \end{tabular}
 \end{table}

The time required to generate a feature vector for an image is about 0.6 milliseconds, the time required to generate clusters for all of the images in an increment is $\sim$1 milliseconds, while pseudo-rehearsal takes an approximate maximum of 14 seconds during the last increment when the system must generate and train on pseudo-exemplars for all of the prior classes. During testing, feature extraction and prediction of an image takes about $\sim$1.2 milliseconds. These results show that our approach has fast training and test times, making it suitable for domestic robot applications when the robot is required to learn and make predictions in near real time.

\begin{figure*}
\centering
\includegraphics[width=0.69\linewidth]{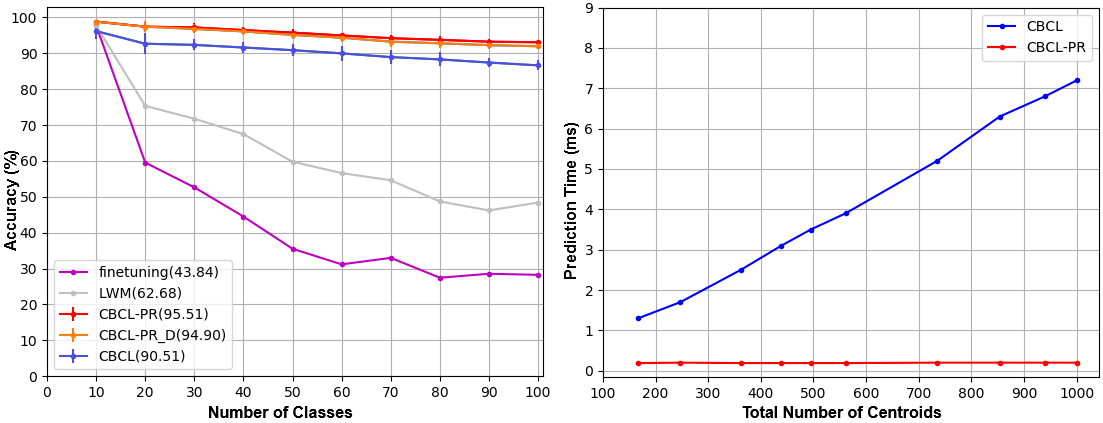}
\caption{\small (Left) Results on the Caltech-101 dataset for continual learning with 10 classes per increment. Mean and standard deviation of the classification accuracy (\%) in each increment over the 10 runs is reported. 
(Right) Comparison of prediction time (in milliseconds) of CBCL and CBCL-PR corresponding to the total number of centroids/clusters learned in 10 increments on the Caltech-101 dataset.}
\label{fig:caltech_cubs}
\end{figure*}  

\subsection{Results on the Caltech-101 Dataset}
\label{sec:Caltech-101_cbcl_pr}
\noindent Figure \ref{fig:caltech_cubs} (left) shows a comparison of CBCL-PR and CBCL-PR\_D to FT, LWM, and CBCL on the Caltech dataset with 10 classes per increment. 
CBCL-PR and CBCL-PR\_D both outperform LWM by $\sim$\textbf{44\%} and $\sim$\textbf{43\%}, respectively, after learning all 100 classes. 
Our methods also outperform CBCL by a margin of $\sim$\textbf{6\%}. Similar to the results on CIFAR-100, FT suffers from catastrophic forgetting and its final classification accuracy is 69.52\% lower than the base accuracy (in the first increment). LWM produces better performance than FT but its final accuracy also decreases by 49.36\% with respect to the base accuracy. In contrast, for CBCL-PR and CBCL-PR\_D the decrease in accuracy is only \textbf{5.8\%} and \textbf{6.84\%}, which indicate that our method does not suffer from catastrophic forgetting.

\subsubsection{FSIL Results}For 5-shot and 10-shot incremental learning, CBCL-PR produces \textbf{88.67\%} and \textbf{91.54\%} average incremental accuracy, respectively. FT and LWM, however, produce only 43.64\% and 62.67\% average incremental accuracy even when they use the complete training set for each class. Thus, CBCL-PR for 5-shot incremental learning produces \textbf{26\%} higher accuracy than LWM, which uses all the training examples for each class. CBCL produces 87.70\% and 89.92\% average incremental accuracy for 5-shot and 10-shot incremental learning, respectively, which is $\sim$1\% lower than CBCL-PR. These results are in accordance with the FSIL results on CIFAR-100.

We also tested FLB on 5-shot and 10-shot incremental learning and compared it with CBCL-PR. For 5-shot and 10-shot incremental learning, FLB produces 72.48\% and 83.81\% average incremental accuracy, respectively. CBCL-PR outperforms FLB with a significant margin (\textbf{16.19\%} and \textbf{7.71\%} lower) for both 5-shot and 10-shot incremental learning. These results reinforce the CIFAR-100 results, demonstrating the value of our method for FSIL. 

\subsubsection{Prediction Time Analysis}
\label{sec:prediction_time_analysis_caltech}
Figure \ref{fig:caltech_cubs} (right) compares the prediction times of CBCL and CBCL-PR corresponding to the total number of centroids/clusters learned in 10 increments on the Caltech-101 dataset. The results clearly show that the prediction time of CBCL-PR is significantly lower than CBCL. Further, the prediction time of CBCL-PR remains constant irrespective of the total number of centroids learned. In contrast, CBCL's prediction time increases drastically with the increase in the number of centroids learned.

\subsection{Ablation Study}
\label{sec:ablation_cbcl_pr}
\noindent We performed an ablation study to better understand CBCL-PR's performance and to quantify the contribution of different stages of processing towards the overall performance. This experiment was performed on the CIFAR-100 dataset with increments of 10 classes and memory budget of $K=$1600 clusters using all the training data per class. Only the diagonal entries of the covariance matrices are stored for this study. We report average incremental accuracy. 
 
We created hybrid versions of CBCL-PR to ablate the clustering and pseudo-rehearsal components. \textit{Trad-Agg-PR} uses traditional agglomerative clustering, \textit{k-means-PR} uses $K$-means clustering to generate centroids and covariance matrices for all the image classes. \textit{CBCL} does not use pseudo-rehearsal. Instead, it directly uses a $k$-Nearest Neighbors--based weighted voting scheme on the centroids (generated by \textit{Agg-Var} clustering) for image classification \cite{Ayub20}. \textit{Trad-Agg} uses traditional agglomerative clustering with a weighted voting scheme for classification (no pseudorehearal). \textit{k-means} uses $K$-means clustering with the weighted voting scheme for classification. Lastly, \textit{NCM} uses an NCM classifier with a pre-trained feature extractor. For each of the hybrids, all other components remained the same as in CBCL-PR.

 \begin{table}
 \caption{Effect on average incremental accuracy by switching off each component separately in CBCL-PR.} 
 \label{tab:Ablation_cbcl_pr}
\centering
\begin{tabular}{ |P{3cm}|P{3.0cm}|}
     \hline
    \textbf{Methods} & \textbf{Accuracy (\%)} \\
     \hline
    Trad-Agg-PR & 64.7\\
    \hline
    k-means-PR & 65.0\\
    \hline
    CBCL & 69.0 \\
     \hline
    Trad-Agg & 59.2\\
    \hline
    k-means & 60.0\\
    \hline
    NCM & 58.2 \\
    \hline
    \textbf{CBCL-PR} & \textbf{73.6} \\
 \hline
 \end{tabular}
 
 \end{table}

Table~\ref{tab:Ablation_cbcl_pr} shows the results of the ablation study. All of the hybrid methods are less accurate than the complete CBCL-PR algorithm. The most drastic decrease in accuracy is seen for \textit{Trad-Agg}, \textit{k-means}, and \textit{NCM} hybrids. Among these three hybrids, \textit{NCM} shows the lowest accuracy because it only uses a single centroid to represent each class and simply uses the closest centroid for the classification of an unlabeled image. Both \textit{Trad-Agg} and \textit{k-means} are significantly inferior to CBCL-PR ($\sim$\textbf{14.0\%}) because they use different clustering techniques and do not use pseudo-rehearsal. \textit{Trad-Agg-PR} and \textit{k-means-PR} achieve higher accuracy than simple \textit{Trad-Agg} and \textit{k-means} showing the significance of pseudo-rehearsal towards the overall accuracy. Particularly, it shows that pseudo-rehearsal technique is effective at regenerating feature vectors that can cover the original feature space which helps improve the accuracy of the classifier. In contrast, for the weighted voting scheme, the centroids simply represent the mean of the feature vectors of the classes and not the entire feature space, which results in the loss of some information and produces inferior accuracy. \textit{CBCL} achieves the highest accuracy among all the hybrids since it uses the \textit{Agg-Var} clustering to generate centroids. Still, its accuracy is lower than CBCL-PR because it does not use pseudo-rehearsal. Note that even CBCL beats the state-of-the-art approaches, which shows that \textit{Agg-Var} clustering is the most important component of our approach. Pseudo-rehearsal's contribution is also significant, increasing the accuracy of all approaches by a margin of $\sim$4.0-5.0\%.

\subsection{FSIL Experiments on a Robot} 
\label{sec:few_shot_incremental_cbcl_pr_robot}
\noindent This experiment evaluates 
a robot's capability to learn to classify new objects continually using CBCL-PR from only a small number of labeled examples (5 or 10) provided by a person. Note, that the only purpose of the robot in this experiment was to use its camera to learn household objects and then classify the test objects. The robot also moved its arm to point to the classified object. Overall, this setup is the same as a traditional FSIL setup applied to real household objects captured from the robot's camera.

\begin{figure}[t]
\centering
\includegraphics[width=1.0\linewidth]{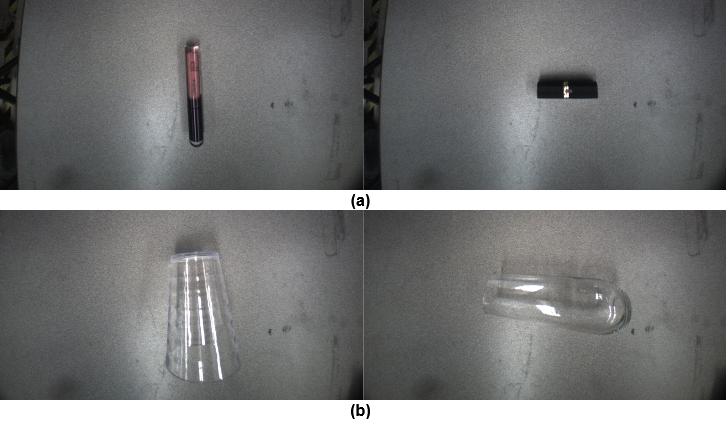}
\caption{\small Some example images of object classes lipstick and glass learned by the robot using the camera in the robot's hand.}
\label{fig:objects}
\end{figure}

We used 22 object classes for the experiment. Each object class was composed of 15 objects. We took pictures of 2 poses of each object, generating 30 images per class. For the 5-shot and 10-shot incremental learning experiments, we used 5 and 10 training images per class. The rest of the images for each class were used for testing. 
We trained the robot incrementally with 2 classes per increment. Example images of the objects used in this experiment are shown in Figure \ref{fig:objects}. Unlike the object images in the benchmark datasets, these images are not ideal. The background of the objects is not perfect and some objects are small in size compared to other objects. 

CBCL-PR was compared to Fine-Tuning (FT), Few-shot Learning Baseline (FLB) and CBCL. We conducted this experiment 10 times with randomized order of classes presented to the robot in different increments. Average accuracies over all ten rounds are presented as results.

Table \ref{tab:few_shot_results_cbcl_pr} compares CBCL-PR against the three methods (FT,  FLB, and CBCL) for 5-shot and 10-shot incremental learning. Similar to the results on benchmark datasets, FT suffers from catastrophic forgetting. CBCL produces good results on both 5-shot and 10-shot learning, achieving accuracy slightly lower than FLB. CBCL-PR, however, produces the best results and even outperforms FLB which does not learn continually. These results depict the effectiveness of our approach for FSIL in real-world conditions using a physical robot.  

During the teaching phase, the time required to collect a single object image by the robot takes only $\sim$2 seconds and the time required to extract features for the image takes about 0.6 milliseconds. In the learning phase, the time required to learn clusters for a new batch of classes takes approximately 1.0 milliseconds and pseudo-rehearsal takes approximately 3 seconds. During the prediction phase, the time required to extract the features and to make a prediction about a test image takes $\sim$0.6 milliseconds.


\begin{table}[t]
\centering
\caption{Comparison of CBCL-PR to FT, FLB and CBCL for FSIL in terms of average incremental accuracy (\%).}
\begin{tabular}{ |P{1.0cm}|P{1.3cm}|P{1.3cm}|P{1.0cm}|P{1.8cm}|}
     \hline
    \textbf{k\_shot} & \textbf{FT} & \textbf{FLB} & \textbf{CBCL} & \textbf{CBCL-PR} \\
     \hline
    5 & 27.16 & 93.58 & 92.97 & \textbf{94.85}\\
     \hline
    10 & 27.36 & 96.76 & 95.67 & \textbf{97.50}\\
 \hline
 \end{tabular}

 \label{tab:few_shot_results_cbcl_pr}
 \end{table}
\section{Conclusion}
\label{sec:conclusion}
\noindent In this paper, we have presented a novel cognitively-inspired framework (CBCL-PR) for a challenging but practical problem: few-shot incremental learning (FSIL). Our approach uses ideas of concept learning to develop computational techniques, such as \textit{Agg-Var} clustering and pseudo-rehearsal to mitigate catastrophic forgetting while learning incrementally from limited data. Extensive experiments on two object classification datasets have shown that CBCL-PR outperforms SOTA methods by significant margins for class-incremental learning and FSIL. We also performed FSIL experiments on a robot for learning household objects from a few examples provided by a human. Our results show that CBCL-PR may have applications for FSIL on real robots. 

One 
limitation of our approach is that it was not tested with real human participants for learning household objects. In the FSIL experiment with a real robot, household objects were taught by the experimenter. In our future work, we plan to run user studies with human participants to continually teach household objects to a robot. Further, the training time required by CBCL-PR is similar to that of a batch learner as both techniques retrain on a similar amount of data. We hope to improve the training time of our approach in the future.

We believe that our approach can be applied in the real world for learning objects continually from a few shots from likely impatient human users that require quick results. We hope that this work will evolve into novel robotics applications, such as domestic cleaning robots and industrial packaging robots that can continually adapt for long-term operation in real-world environments.

\section*{Acknowledgments}
\noindent 
Preliminary version of this work was presented in \cite{Ayub20,Ayub_2020_BMVC}. This work significantly extends these works in several ways. First we have updated \textit{Agg-Var} clustering and introduced a memory consolidation (pseudo-rehearsal) phase to improve on our previous approach (termed CBCL). Second, we have added a prediction time analysis of CBCL to show its limitations. We have also performed experiments on the Caltech-101 dataset to show the improvement in prediction time by CBCL-PR. Third, we have performed new experiments using CBCL-PR on two datasets and on a real robot. We have also added comparison of CBCL-PR with most recent FSIL methods in our experiments. Finally, the paper contains a more complete background and related work, and provides a more detailed explanation of the entire architecture through various intuitive examples.

This  work  was  partially  supported  by  the  Air  Force  Office  of  Sponsored  Re-search contract FA9550-17-1-0017 and National Science Foundation grant CNS-1830390. Any opinions, findings, and conclusions or recommendations expressed in this material are those of the author(s) and do not necessarily reflect the views of the National Science Foundation.

%

{\small
\bibliographystyle{IEEEtran}
\bibliography{main}
}

{\appendices
\section{Cluster Reduction}
\label{sec:cluster_reduction}
\noindent
Implementations on real hardware only have access to limited memory (i.e. disk space). It is thus important to consider the memory footprint required for a continual learning algorithm~\cite{Rebuffi_2017_CVPR}. We therefore propose a novel method that restricts the number of clusters while attempting to maintain classification accuracy. 

Assume that a system can store a maximum of $K$ clusters (centroids and covariance matrices) and that it currently has stored $K_t$ clusters for $t$ classes. For the next batch of classes the system needs to store $K_{new}$ more clusters but the total number of clusters is $K_t + K_{new} > K$. Hence, the system needs to reduce the total stored clusters to $K_r = K_t+K_{new}-K$. Rather than reducing the number of clusters for each class equally, CBCL-PR reduces the clusters for each class proportional to the number of clusters for the class before reduction. The reduction in the number of clusters  $N_y^{'}$ for each class $y$ is calculated as (whole number):
\begin{equation}
    N^{'}_{y}(new) = N_y^{'}(1-\frac{K_r}{K_t})
\end{equation}

\noindent where $N_y^{'}(new)$ is the number of clusters for class $y$ after reduction. Rather than simply removing the extra clusters from each class, we cluster the closest clusters in each class to get new clusters, keeping as much information as possible about the previous classes. This process is accomplished by applying k-means clustering~\cite{jain99} on the cluster set of each class $y$ to reduce them to a total of $N_y^{'}(new)$ clusters. 

\section{Analysis of Different Memory Budgets on CIFAR-100}
\label{sec:mem_budget_cbcl_pr}
\noindent We perform a set of experiments on the CIFAR-100 dataset to analyze the effect of different memory budgets on the performance of CBCL-PR. We only use the diagonal entries of covariance matrices for this set of experiments. To show the significance of our cluster reduction technique, we also show results when clusters are simply removed when the memory limit is reached. Figure \ref{fig:memory_budgets_cbcl_pr} compares the average incremental accuracy of CBCL-PR with our proposed cluster reduction technique and with simple removal of clusters for different memory budgets. As expected, both approaches achieve higher accuracy when provided with higher memory budgets. However, CBCL-PR using the cluster reduction technique constantly outperforms CBCL-PR with simple removal of clusters for all memory budgets (except for $K=1600$ when there is no need for any reduction), and the performance gap increases for smaller memory budgets. This clearly shows the effectiveness of our proposed cluster reduction technique over simple removal of clusters. Furthermore, it should be noted that even for only $K=1000$ clusters, CBCL-PR's average incremental accuracy (\textbf{65.4\%}) is higher than that of the state-of-the-art methods. 

\begin{figure}[t]
\centering
\includegraphics[width=1.0\linewidth]{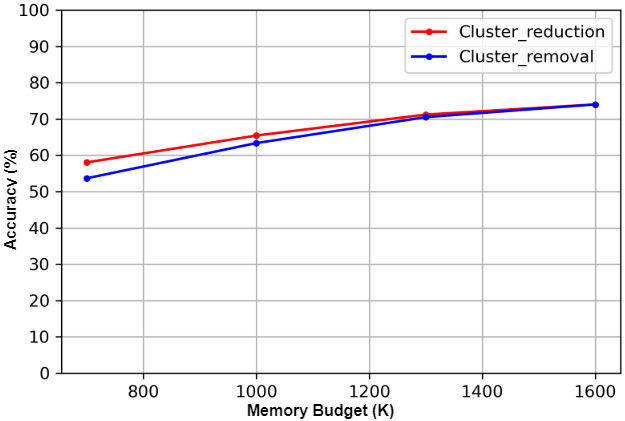}
\caption{\small Average incremental accuracy of CBCL-PR with the centroid reduction technique and centroid removal for different memory budgets (K: total number of clusters).}
\label{fig:memory_budgets_cbcl_pr}
\end{figure}

\begin{figure}[t]
\centering
\includegraphics[width=1.0\linewidth]{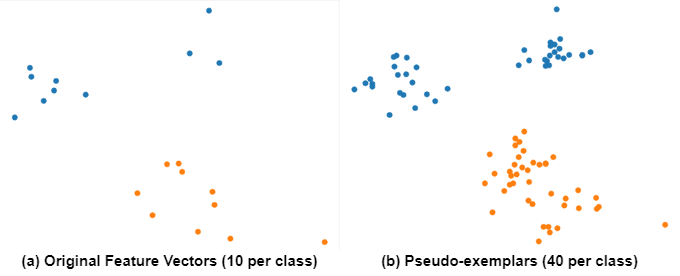}
\caption{\small Comparison of the original feature space (10 feature vectors per class) with 80 pseudo-exemplars generated by 6 clusters (3 per class) for 2 classes in the MNIST dataset.}
\label{fig:encoded_ep}
\end{figure}

\section{Illustration of Pseudorehearsal for FSIL}
An argument can be made that applying pseudorehearsal to Gaussian distributions learned through FSIL could be vulnerable. It could lead to unexpected features and, thus, negative classification performance. To test this, we showed on CIFAR-100 dataset (Table II in the main paper) that pseudorehearsal actually improves the overall accuracy instead of effecting it negatively. To further illustrate this, we performed a similar experiment as in Section III-B with 2 classes of the MNIST dataset. However, instead of using all the feature vectors, we used only 10 feature vectors per class (Figure \ref{fig:encoded_ep} (left)). The 2D space covered by the 40 pseudo-exemplars per class looks similar to the original feature space, although it is more dense than the original (Figure \ref{fig:encoded_ep} (right)). This shows that CBCL-PR is able to generate more data covering the original feature space to help improve the classification performance.

}

\end{document}